%% file: main.tex
\documentclass[10pt,twocolumn,letterpaper]{article}

\usepackage[pagenumbers]{cvpr} %

\input{preamble}

\definecolor{cvprblue}{rgb}{0.21,0.49,0.74}
\usepackage[pagebackref,breaklinks,colorlinks,allcolors=cvprblue]{hyperref}

\def\paperID{7} %
\def\confName{CVPR}
\def\confYear{2026}

\title{Towards Data-Efficient Video Pre-training with Frozen Image Foundation Models}

\author{
Svetlana Orlova
\quad Niccolò Cavagnero
\quad Gijs Dubbelman
\\
Eindhoven University of Technology\\
{\tt\small s.orlova@tue.nl}
}

\begin{document}
\maketitle
\input{sec/0_abstract}
\input{sec/1_intro}

\input{sec/2_relwork}

\input{sec/31_preliminaries}

\input{sec/32_method}
\input{sec/41_setup}
\input{sec/42_experiments_v2}
\input{sec/5_conclusion}
\input{sec/6_acknowledgements}
{
    \small
    \bibliographystyle{ieeenat_fullname}
    \bibliography{main}
}

\input{sec/X_suppl}

\end{document}

%% file: preamble.tex
\usepackage{multirow}

\newcommand{\gain}[1]{{\color{green!60!black}\scriptsize{+#1}}}    %
\newcommand{\ngain}[1]{{\color{green!60!black}\scriptsize{-#1}}}   %

\usepackage{colortbl}

\newcommand{\mypara}[1]{%
  \textbf{#1}\quad%
}

%% file: sec/0_abstract.tex
\begin{abstract}
Video foundation models achieve strong performance across many video understanding tasks, but typically require large-scale pre-training on massive video datasets, resulting in substantial data and compute costs. In contrast, modern image foundation models already provide powerful spatial representations. This raises an important question: can competitive video models be built by reusing these spatial representations and pre-training only for temporal reasoning?
We take initial steps toward exploring a lightweight training paradigm that freezes a pre-trained image foundation model and trains only a recurrent temporal module to process streaming video. By reusing an image foundation model as a spatial encoder, this approach could significantly reduce the amount of video data and compute required compared to end-to-end video pre-training. In this work, we explore the feasibility of this approach before investing in computing for video pre-training.  
Our empirical findings across multiple video understanding tasks suggest that strong temporal performance can emerge without large-scale video pre-training, motivating future work on recurrent video foundation models obtained by pre-training a temporal module on top of a frozen image foundation model. Code: \href{https://github.com/tue-mps/towards-video-image-frozen}{https://github.com/tue-mps/towards-video-image-frozen}.

\end{abstract}

%% file: sec/1_intro.tex
\section{Introduction}
\label{sec:intro}

\begin{figure}[t]
    \centering
    \includegraphics[width=0.85\linewidth]{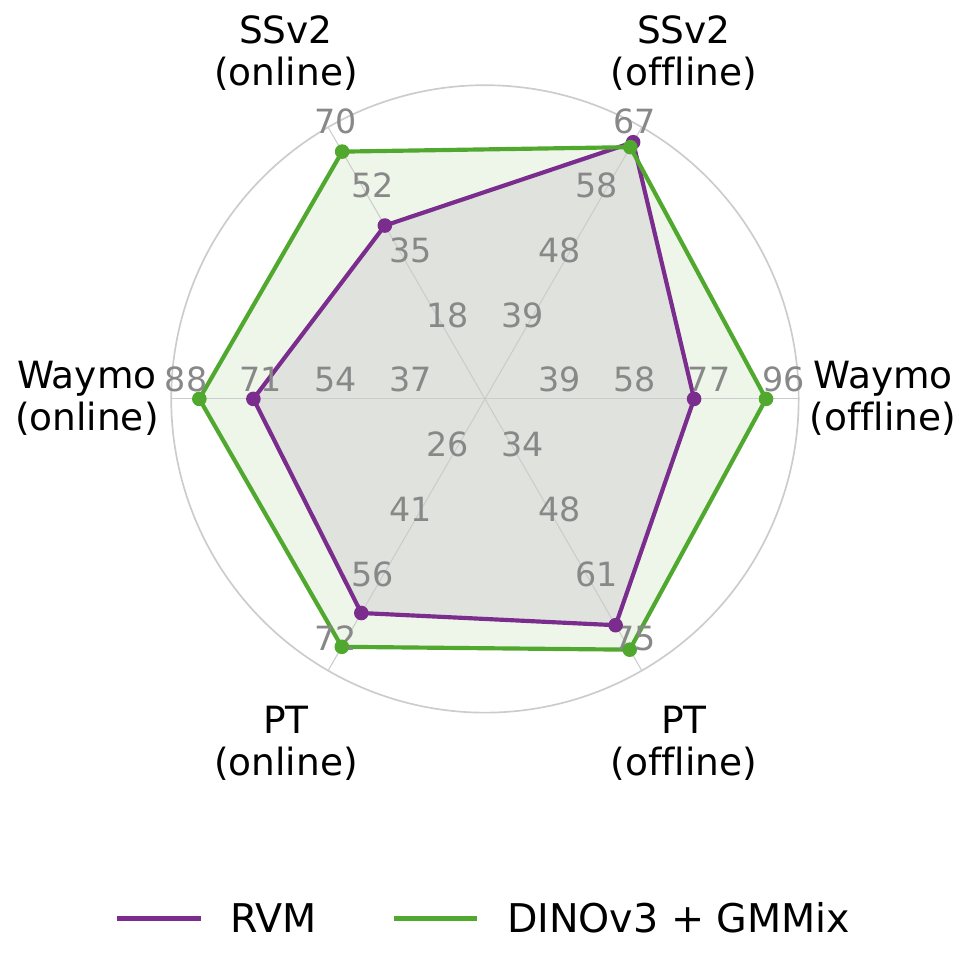}
    \caption{\textbf{Video Foundation Model \vs Image Foundation Model + Recurrent Head.} Comparison of a frozen Video Foundation Model (RVM~\cite{rvm}) \vs a frozen Image Foundation Model (DINOv3~\cite{dinov3}) with a fine-tuned recurrent temporal head, GatedMambaMix (GMMix). DINOv3 achieves similar performance across different tasks without large scale video pre-training.}
    \label{fig:teaser}
\end{figure}

Recent years have seen rapid progress in video foundation models, which aim to learn general-purpose representations for a wide range of video understanding tasks~\cite{videomae,vjepa,4ds,rvm}. Most competitive models adopt large architectures that are massively pre-trained end-to-end on extremely large video datasets, often comprising millions to billions of clips. While this pre-training is required for strong performance, it comes with substantial costs in terms of data collection, storage, and computational resources.

At the same time, image foundation models have reached an unprecedented level of capability. Trained on billions of images, these models provide powerful spatial representations that transfer well across tasks and domains~\cite{oquab_dinov2_2023, dinov3,tschannen2025siglip}. This raises an important question: \textit{to what extent is large-scale video pre-training actually necessary if strong spatial representations are readily available?}

A promising alternative is to leverage a pre-trained image backbone and focus the more computationally intensive video training exclusively on temporal modeling. Instead of learning spatial and temporal representations jointly from scratch on video data, a model could inherit spatial knowledge from an image foundation model and learn temporal reasoning via video pre-training of a temporal module processing the image foundation model's representations. Such a strategy could dramatically reduce both the data and compute requirements needed to develop powerful video models.

In this work, we explore this idea in the context of recurrent video models. We do not yet perform video pre-training of the temporal module, but, in this work, we explore the feasibility of this approach before making the required compute investment. For this, we aim to address two research questions: \textit{(1) is image pre-training of the spatial encoder competitive with video pre-training?, and (2) do we actually need large-scale video pre-training for the temporal module?} 

To answer these questions, we conduct experiments using different image foundation models and temporal architectures. Our evaluation spans several representative tasks: action recognition (Something-Something v2~\cite{ssv2}), object tracking (Waymo Open~\cite{waymo}), point tracking (Perception Test~\cite{perceptiontest}), depth estimation (ScanNet~\cite{scannet}), and camera pose estimation (NuScenes~\cite{nuscenes}), allowing us to comprehensively assess whether strong temporal reasoning can emerge without large-scale video pre-training. The results show that spatial representations obtained from frozen image foundation models are stronger than those obtained with video pre-training, but at the same time, the results also indicate that video pre-training of the temporal module (not yet done in this work) is likely needed to surpass current SotA video foundation models across all settings and tasks.  

This paper should therefore be viewed as a work in progress toward a new pre-training paradigm for recurrent video models. Our goal is to provide initial empirical evidence and insights that inform future efforts to scale this approach into a full video foundation model.

We make the following contributions:
\begin{enumerate}
\item Empirical study of temporal learning with frozen spatial representations.
Across multiple image encoders, temporal architectures, and diverse video tasks, we demonstrate that effective temporal reasoning emerges even with a fixed image-level backbone.

\item Evidence towards data-efficient video pre-training.
Our results indicate that modern image foundation models already contain much of the spatial capacity needed for video understanding, thereby supporting a future paradigm in which only the temporal module requires video pre‑training.
\end{enumerate}

\subsection{Small Data Statement}
The end goal of our line of research is to reduce the data and computational requirements for pre-training recurrent video foundation models. We explore decoupling the pre-training of the spatial (image) encoder from that of the temporal module. By using an off-the-shelf image foundation model as a spatial encoder, we hypothesize that less video data is needed to pre-train the temporal module. This work reports our current findings, which do not yet include temporal pre-training, but it verifies that our paradigm is promising and that investing in temporal pre-training is a promising direction for obtaining more data-efficient pre-training strategies for recurrent video foundation models.

%% file: sec/2_relwork.tex
\section{Related Work}
\label{sec:related}

\subsection{Image Foundation Models}

Vision Transformers~\cite{dosovitskiy2020vit} trained with self-supervised objectives have become the dominant paradigm for visual representation learning.
Contrastive methods such as CLIP~\cite{radford2021clip} learn aligned image-text representations, while masked autoencoders~\cite{mae} learn directly from pixels via reconstruction.
Self-distillation approaches have proven particularly effective as frozen feature extractors: DINOv2~\cite{oquab_dinov2_2023} produces general-purpose features that achieve strong results across classification, segmentation, and depth estimation with only task-specific heads trained on top.
DINOv3~\cite{dinov3} extends this further, establishing a single frozen ViT as a universal vision backbone that sets state-of-the-art results on multiple vision tasks, without any task-specific fine-tuning of the encoder.

We use such image foundation models as frozen spatial feature extractors and investigate whether adding a learned temporal module on top is sufficient to match video foundation models.

\subsection{Video Foundation Models}
\mypara{Video Vision Transformers.}
The dominant approach to video modeling extends ViT by tokenizing video clips into spatio-temporal patches and using pre-training with self-supervised objectives on large-scale video data.
VideoMAE~\cite{videomae} applies tube masking to spatio-temporal patches, V-JEPA~\cite{vjepa} predicts masked representations in latent space, and 4DS~\cite{4ds} simplifies the objective and focuses on scaling, showing consistent improvement on geometric and temporal tasks up to 22B parameters.
These architectures process fixed-length clips and are inherently non-causal, requiring the entire temporal window upfront. All are trained end-to-end on billions of video samples, with 4DS demonstrating that performance continues to improve with further scaling.

\mypara{Recurrent Architectures.}
An alternative direction explores recurrent architectures designed specifically for sequential video processing.
Selective state space models~\cite{mamba} offer expressive state evolution while scaling linearly with sequence length. VideoMamba~\cite{li2024videomamba} applies them to video, typically as full backbone architectures with bidirectional processing.
TRecViT~\cite{patraucean2024trecvit} factorizes video modeling into time-space-channel dimensions with linear recurrent units for temporal mixing, achieving competitive results with causal processing.
RVM~\cite{rvm} factorizes the model into spatial and temporal parts, with the architecture consisting of a ViT followed by a GRU-gated recurrent core, and trains both components jointly on billions of video samples.
While competitive, they generally underperform compared to video foundation models based on Video ViT, which have direct access to all temporal information in the encoder and are pre-trained at significantly larger scale.
A notable exception is RVM, which demonstrates competitive and often stronger video understanding capabilities than plain ViT-based video architectures~\cite{rvm}. Therefore, we consider RVM as our baseline recurrent video foundation model for our work.  
RVM is pre-trained end-to-end with a video-masked autoencoder objective on approximately 8.4M video clips.

The end-to-end pre-training employed by current video foundation models is extremely data- and compute-intensive, and in this work, we explore whether similar or even better video understanding can be achieved by reusing the strong spatial understanding of an image foundation model and only training a temporal module. Ultimately, such an approach would significantly reduce the data and compute needed for video models.

%% file: sec/31_preliminaries.tex
\section{Methodology}
\label{sec:method}

\subsection{Preliminaries}
\label{sec:preliminaries}

\mypara{Vision Transformer.}
The Vision Transformer (ViT)~\cite{dosovitskiy2020vit} divides an image $I \in \mathbb{R}^{H \times W \times 3}$ into $N$ non-overlapping patches of shape $p \times p$, which are linearly projected into patch tokens $\mathbf{X}^0 \in \mathbb{R}^{N \times D}$ and processed by $L$ transformer blocks.
Each block $i$ applies multi-head self-attention (MHSA) and a two-layer MLP with a non-linear activation:
\begin{align}
    \mathbf{Z}^i &= \mathbf{X}^i + \mathrm{MHSA}(\mathrm{LN}(\mathbf{X}^i)), \label{eq:vit_mhsa}\\
    \mathbf{X}^{i+1} &= \mathbf{Z}^i + \mathrm{MLP}(\mathrm{LN}(\mathbf{Z}^i)), \label{eq:vit_mlp}
\end{align}
where $\mathrm{LN}$ is layer normalization~\cite{ba2016layer}.
The final patch tokens $\mathbf{X}^L \in \mathbb{R}^{N \times D}$ serve as spatial features, where $N = \frac{HW}{p^2}$ is the number of patches.
Both image and video foundation models are built on this base architecture.

\mypara{Recurrent Video Masked Autoencoders (RVM).}
RVM~\cite{rvm} separates spatial encoding from temporal processing.
Given a frame $I_t$ from a video sequence, a ViT encoder $\mathcal{E}$ extracts per-frame features $\mathbf{X}^L_t = \mathcal{E}(I_t) \in \mathbb{R}^{N \times D}$, a recurrent temporal module $\mathcal{S}$ updates a hidden state $\mathbf{h}_t, \mathbf{s}_t = \mathcal{S}(\mathbf{X}^L_t, \mathbf{s}_{t-1})$, and a task-specific readout $\mathcal{R}$ produces predictions $\hat{y}_t = \mathcal{R}(\mathbf{h}_t)$.
Despite this architectural separation, RVM trains both the encoder and the temporal module jointly end-to-end via asymmetric masked prediction on ${\sim}$8.4M video clips with 95\% masking and $L_2$ pixel reconstruction loss.

\mypara{RVM\textsubscript{RNN}.}
RVM's temporal module, hereafter RVM\textsubscript{RNN}, combines GRU-style gating~\cite{gru} with a cross-attention transformer:
\begin{align}
    \mathbf{u}_t &= \sigma\!\left(\mathbf{W}^u_f \mathbf{X}^L_t + \mathbf{W}^u_s \mathbf{s}_{t-1}\right), \label{eq:rvmrnn_update}\\
    \mathbf{r}_t &= \sigma\!\left(\mathbf{W}^r_f \mathbf{X}^L_t + \mathbf{W}^r_s \mathbf{s}_{t-1}\right), \label{eq:rvmrnn_reset}\\
    \tilde{\mathbf{h}}_t &= \mathrm{Tx}\!\left(\mathbf{X}^L_t,\; \mathbf{r}_t \odot \mathrm{LN}(\mathbf{s}_{t-1})\right), \label{eq:rvmrnn_candidate}\\
    \mathbf{s}_t &= (1 - \mathbf{u}_t) \odot \mathbf{s}_{t-1} + \mathbf{u}_t \odot \tilde{\mathbf{h}}_t, \label{eq:rvmrnn_state}
\end{align}
where $\mathbf{u}_t$ and $\mathbf{r}_t$ are update and reset gates, $\mathrm{Tx}$ is a transformer consisting of $K$ layers, each fusing the current frame features with the gated state via cross-attention followed by an MLP and self-attention, and the state $\mathbf{s}_t \in \mathbb{R}^{N \times D}$ maintains one vector per spatial token.
The output is $\mathbf{h}_t = \mathrm{LN}(\mathbf{s}_t)$.

\mypara{Mamba.}
Mamba~\cite{mamba} is a selective state space model (SSM). A classical SSM maps an input sequence to an output sequence through a latent state:
\begin{align}
    \mathbf{h}_t &= \mathbf{A}\,\mathbf{h}_{t-1} + \mathbf{B}\,x_t, \label{eq:ssm_state}\\
    y_t &= \mathbf{C}\,\mathbf{h}_t, \label{eq:ssm_output}
\end{align}
where $\mathbf{A}$, $\mathbf{B}$, $\mathbf{C}$ are fixed matrices.
Mamba makes these matrices input-dependent, $\mathbf{B}_t = \mathbf{B}(x_t)$, $\mathbf{C}_t = \mathbf{C}(x_t)$, allowing the model to selectively retain or discard information based on the current input.
This provides a recurrent framework that naturally supports causal processing and scales linearly with sequence length, while maintaining efficient training through its parallelizable recurrence formulation.

%% file: sec/32_method.tex
\subsection{Framework}
\label{sec:framework}

Our framework decouples spatial and temporal learning for video understanding.
It consists of three components: a frozen image encoder that provides spatial features, a recurrent temporal module that builds temporal representations causally, and an attentive readout head that produces task-specific predictions.

Given a video $V = \{I_1, \ldots, I_T\}$ with $T$ frames, our model processes it in three stages.

\mypara{Frozen Image Encoder.}
Each frame $I_t \in \mathbb{R}^{H \times W \times 3}$ is independently processed by a frozen pre-trained image encoder $\mathcal{E}$.
The encoder is kept completely frozen; no gradients flow through it during training.
We primarily use DINOv3~\cite{dinov3} as our image encoder, though the framework is encoder-agnostic and we evaluate multiple encoders with different pre-training objectives in \cref{sec:experiments}.
By default, we use multi-depth feature extraction as described below.

\mypara{Multi-depth Feature Extraction.}
When an encoder is fine-tuned end-to-end, it can learn to consolidate task-relevant information into its final layer.
A frozen encoder, however, retains useful spatial information distributed across its depth: early layers capture low-level structure, while deeper layers encode higher-level semantics.
To exploit this, we extract patch tokens $\mathbf{F}_{t,j} \in \mathbb{R}^{N \times D}$ from four equally spaced ViT depths ($j = 1, \ldots, 4$; at relative depths $1/4$, $1/2$, $3/4$, and $1$).
Each layer's features are adapted by a trainable per-layer MLP with a residual connection, and the final representation is the mean across depths:
\begin{align}
    \hat{\mathbf{F}}_{t,j} &= \mathbf{F}_{t,j} + \mathrm{MLP}_j(\mathrm{BN}(\mathbf{F}_{t,j})), \label{eq:interm_mlp}\\
    \mathbf{X}_t &= \frac{1}{4} \sum_{j=1}^{4} \hat{\mathbf{F}}_{t,j}. \label{eq:interm_avg}
\end{align}
The \texttt{CLS} and register tokens from the final encoder layer are concatenated with $\mathbf{X}_t$ to form the input token sequence passed to the temporal module.
This provides richer multi-scale spatial information than final-layer features alone, and we show in \cref{sec:experiments} that this consistently improves performance across all temporal architectures.

\mypara{Recurrent Temporal Module.}
The per-frame features are processed sequentially by a recurrent temporal module $\mathcal{S}$ that maintains a hidden state across frames:
\begin{equation}
    \mathbf{h}_t, \mathbf{s}_t = \mathcal{S}(\mathbf{X}_t, \mathbf{s}_{t-1}),
    \label{eq:seqcore}
\end{equation}
where $\mathbf{s}_t$ is the recurrent state and $\mathbf{h}_t \in \mathbb{R}^{N \times D'}$ is the output representation for frame $t$.
The state is initialized to zeros: $\mathbf{s}_0 = \mathbf{0}$.
The temporal module processes frames causally, it never accesses future frames.
It is trained from scratch alongside the readout, while the encoder remains frozen.
In \cref{sec:temporal}, we detail the different recurrent temporal models used for our research.

\mypara{Attentive Readout.}
To isolate the contribution of the recurrent temporal module, we employ a streaming protocol where the readout receives only the current frame's output $\mathbf{h}_t$:
\begin{equation}
    \hat{y}_t = \mathcal{R}_{\mathrm{stream}}(\mathbf{h}_t), \qquad t = 1, \ldots, T.
    \label{eq:streaming}
\end{equation}
The readout architectures are based on~\cite{rvm,4ds}, but operate on $N$ tokens instead of $T \times N$.
All temporal context must therefore reside in the recurrent state $\mathbf{s}_t$: if the temporal module fails to accumulate useful information, streaming predictions degrade since the readout cannot compensate.

For video-level tasks, the streaming model must produce a single prediction from the per-frame outputs.
For action recognition (SSv2), we use only the last frame's prediction $\hat{y}_T$, since it benefits from the full accumulated temporal context.
For frame-level tasks, the readout produces a prediction independently at each time step, attending only to the $N$ spatial tokens of that frame.

We additionally compare our models to state-of-the-art video foundation models following the offline evaluation protocol of~\cite{rvm,4ds}, where the readout attends over all frames simultaneously, enabling fair comparison with baselines that report results under this protocol. Our streaming evaluation (\cref{eq:streaming}) provides the stricter test of temporal representations, as it prevents the readout from compensating for temporal modeling.

\mypara{Evaluation Strategy.}
The modular design of our framework enables controlled experiments along several axes.
To address the first research question from \cref{sec:intro}, we compare frozen encoders with different pre-training paradigms (image vs.\ video).
To address the second, we leverage RVM's decoupled architecture to initialize our temporal module with video pre-trained weights and measure the benefit over training from scratch.
Additionally, we evaluate multiple temporal architectures to assess whether a specific design is critical, and compare multi-depth and final-layer feature extraction strategies.

\subsection{Temporal Module Architectures}
\label{sec:temporal}

We investigate four temporal architectures.
Alongside RVM's Gated Transformer Core~\cite{rvm}, we use the default Mamba block~\cite{mamba} as a lightweight baseline, and introduce two novel extensions that progressively incorporate the architectural inductive biases of RVM\textsubscript{RNN}: MambaMix adds spatial self-attention within each frame, and GMMix further incorporates gated temporal updates, making it the closest Mamba-based analogue to RVM\textsubscript{RNN}.
All share the same recurrent interface (\cref{eq:seqcore}), making them interchangeable within our framework.

\mypara{RVM\textsubscript{RNN}.}
We adopt the recurrent core from RVM~\cite{rvm} (\cref{eq:rvmrnn_update,eq:rvmrnn_state}) as our first temporal architecture, denoted RVM\textsubscript{RNN}.
This is the most structurally complex design among our four options, combining GRU gating, spatial self-attention, and cross-attention within a single module.

\mypara{Mamba.}
Our simplest module applies a selective SSM (\cf \cref{sec:preliminaries}) with a pre-norm residual connection, operating independently on each spatial token across time:
\begin{equation}
    \mathbf{x}^{k+1} = \mathbf{x}^k + \mathrm{Mamba}\!\left(\mathrm{LN}(\mathbf{x}^k)\right).
    \label{eq:mamba_residual}
\end{equation}
This layer is repeated $K$ times, and the output is $\mathbf{h}_t = \mathrm{LN}(\mathbf{x}^K_t)$.
This design provides no cross-patch spatial interaction; each patch evolves independently.

\mypara{MambaMix.}
To introduce spatial reasoning, each MambaMix layer interleaves two operations:
\begin{align}
    \mathbf{z}^k &= \mathrm{SpatialBlock}\!\left(\mathbf{x}^k\right), \label{eq:mambamix_spatial}\\
    \mathbf{x}^{k+1} &= \mathbf{z}^k + \mathrm{Mamba}\!\left(\mathrm{LN}(\mathbf{z}^k)\right), \label{eq:mambamix_temporal}
\end{align}
where the $\mathrm{SpatialBlock}$ applies self-attention and an MLP across all $N$ patches within each frame independently, and the temporal Mamba processes each patch across $T$ frames as in \cref{eq:mamba_residual}.
This is repeated for $K$ layers, and the output is $\mathbf{h}_t = \mathrm{LN}(\mathbf{x}^K_t)$.
MambaMix adds cross-patch spatial context before temporal processing, allowing patches to share information within each frame.

\mypara{GatedMambaMix (GMMix).}
GMMix extends MambaMix with a learned gating mechanism that controls how much temporal information to incorporate.
After the spatial block and temporal Mamba, a gate interpolates between the pre- and post-Mamba representations:
\begin{align}
    \mathbf{z}^k &= \mathrm{SpatialBlock}\!\left(\mathbf{x}^k\right), \label{eq:gmmix_spatial}\\
    \tilde{\mathbf{z}}^k &= \mathbf{z}^k + \mathrm{Mamba}\!\left(\mathrm{LN}(\mathbf{z}^k)\right), \label{eq:gmmix_mamba}\\
    \mathbf{g}^k &= \sigma\!\left(\mathrm{Gate}([\mathbf{z}^k;\, \tilde{\mathbf{z}}^k])\right), \label{eq:gmmix_gate}\\
    \mathbf{x}^{k+1} &= (1 - \mathbf{g}^k) \odot \mathbf{z}^k + \mathbf{g}^k \odot \tilde{\mathbf{z}}^k, \label{eq:gmmix_blend}
\end{align}
where $[\cdot\,;\,\cdot]$ denotes concatenation along the feature dimension and $\mathrm{Gate}$ is a linear projection from $2D$ to $D$, applied independently to each token.
This is repeated for $K$ layers, and the output is $\mathbf{h}_t = \mathrm{LN}(\mathbf{x}^K_t)$.
The gate provides explicit control over temporal information flow, analogous to the GRU gating in RVM\textsubscript{RNN} (\cref{eq:rvmrnn_state}).

%% file: sec/41_setup.tex
\begin{figure*}[t]
    \centering
    \includegraphics[width=0.95\linewidth]{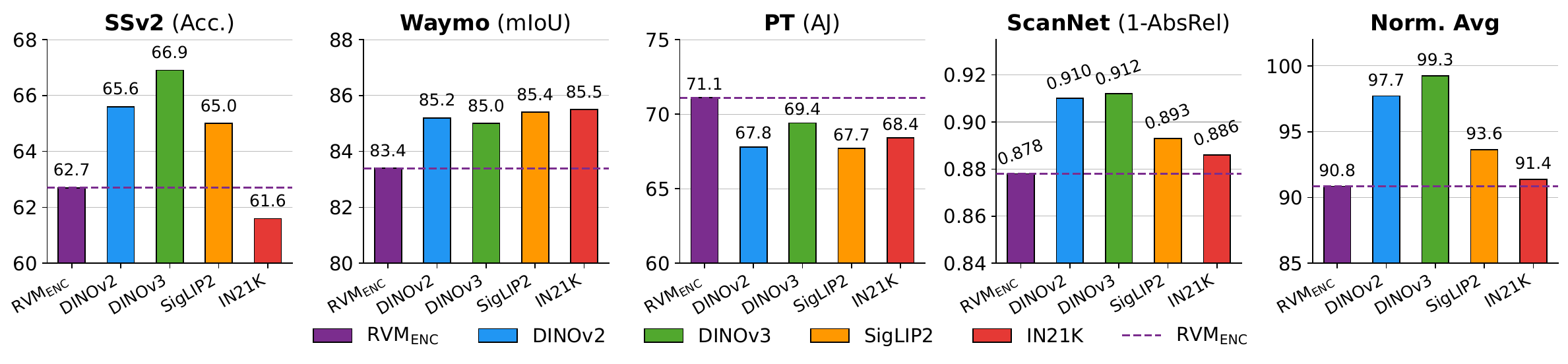}
    \caption{\textbf{Image Pre-training \vs Video Pre-training.} GMMix temporal module paired with various pre-trained encoders. All encoders are frozen, only GMMix and the readout are trained from scratch. Image pre-trained encoders consistently match or outperform the video pre-trained RVM encoder.}
    \label{fig:encoder_comparison}
\end{figure*}

\input{tables/4_streaming}

\section{Experiments}
\label{sec:experiments}

\subsection{Experimental Setup}
\label{sec:setup}

\mypara{Benchmarks and Protocol.}
We follow the evaluation protocol of~\cite{rvm,4ds} and evaluate on video understanding tasks spanning action recognition (Something-Something v2~\cite{ssv2}, top-1 accuracy), object tracking (Waymo Open~\cite{waymo}, mIoU), and point tracking (Perception Test~\cite{perceptiontest}, Average Jaccard).
Note that for point tracking, all models are trained on the synthetic Kubric MOVi-E dataset~\cite{greff2022kubric} and evaluated on real videos from Perception Test, making this a synthetic-to-real transfer evaluation.
For streaming evaluation, we additionally include depth estimation on ScanNet~\cite{scannet} (AbsRel) and camera pose estimation on NuScenes~\cite{nuscenes} (translational and rotational relative errors).
The pre-trained backbone is frozen and a task-specific cross-attention readout head is trained on top. For our models, both the temporal module and the readout are trained from scratch.

Unless stated otherwise, all experiments use the streaming protocol (\cref{eq:streaming}), where the readout receives only the current frame's tokens. This provides a stricter test of learned temporal representations, as the model cannot rely on the readout to compensate for weak temporal modeling. To compare with video foundation models that report results under the standard multi-frame protocol, we additionally evaluate in the offline setting.

Since no public implementation is available for the evaluation pipelines of~\cite{rvm,4ds}, we re-implement all training and evaluation following the original description.
Full training and evaluation details are provided in the supplementary material.

\mypara{Baselines.}
We compare against spatio-temporal video foundation models (VideoMAE~\cite{videomae}, V-JEPA~\cite{vjepa}, 4DS~\cite{4ds}) and image foundation models (DINOv2~\cite{oquab_dinov2_2023}, DINOv3~\cite{dinov3}) for reference.
Our primary comparison is with RVM~\cite{rvm}, which follows the same decoupled approach of a per-frame encoder and a recurrent temporal core, but trains both components jointly on video data.
In the main comparison, we use \emph{RVM\,(frozen)}, where the full pre-trained model is frozen and only the readout is trained.
Since RVM's architecture is decoupled, we can split it into encoder and temporal core and recombine them with our components, which we exploit in ablations to isolate the contributions of the video pre-trained encoder and temporal module (\cref{sec:ablations}).

\mypara{Normalized Average.}
To aggregate across tasks, we report a normalized average following~\cite{rvm}: each score is divided by the column-best, and the ratios are averaged.

%% file: tables/4_streaming.tex
\begin{table*}
    \centering
    \begin{tabular}{lccccccc}
    \toprule
    Model & Size(M) & SSv2 & Waymo & PT & ScanNet & NuScenes & Norm. Avg \\
    & & Acc. ($\uparrow$\%) & mIoU ($\uparrow$) & AJ ($\uparrow$) & AbsRel ($\downarrow$) & RPE\textsubscript{tr} ($\downarrow$) & ($\uparrow$) \\
    \midrule
    RVM-L      & 375 & 46.9 & 72.7 & 61.3 & 0.1293 & 36.00 & 77.7 \\
    DINOv3-L + RVM\textsubscript{RNN} & 375 & \textbf{67.1} & \textbf{85.7} & 63.7 & 0.0900 & 29.37 & 96.8 \\
    DINOv3-L + M        & 347 & 63.3 & 84.8 & 65.4 & 0.0963 & 28.48 & 95.3 \\
    DINOv3-L + MMix     & 397 & 66.4 & \underline{85.0} & \underline{66.7} & \textbf{0.0870} & \underline{28.13} & \underline{98.8} \\
    DINOv3-L + GMMix    & 405 & \underline{66.9} & \underline{85.0} & \textbf{69.4} & \underline{0.0885} & \textbf{28.09} & \textbf{99.4} \\
    \bottomrule
    \end{tabular}

    \caption{\textbf{Temporal Modules Comparison.} Four temporal modules paired with a frozen DINOv3-L encoder. RVM\textsubscript{RNN} = RVM's recurrent core, M = Mamba, MMix = MambaMix, GMMix = GatedMambaMix. RVM shown as baseline. NuScenes RPE\textsubscript{tr} is in mm. RPE\textsubscript{rot} is similar (0.08--0.09$^\circ$) across all models and omitted.}
    \label{tab:streaming}
\end{table*}

%% file: sec/42_experiments_v2.tex
\begin{figure*}[t]
    \centering
    \includegraphics[width=0.8\linewidth]{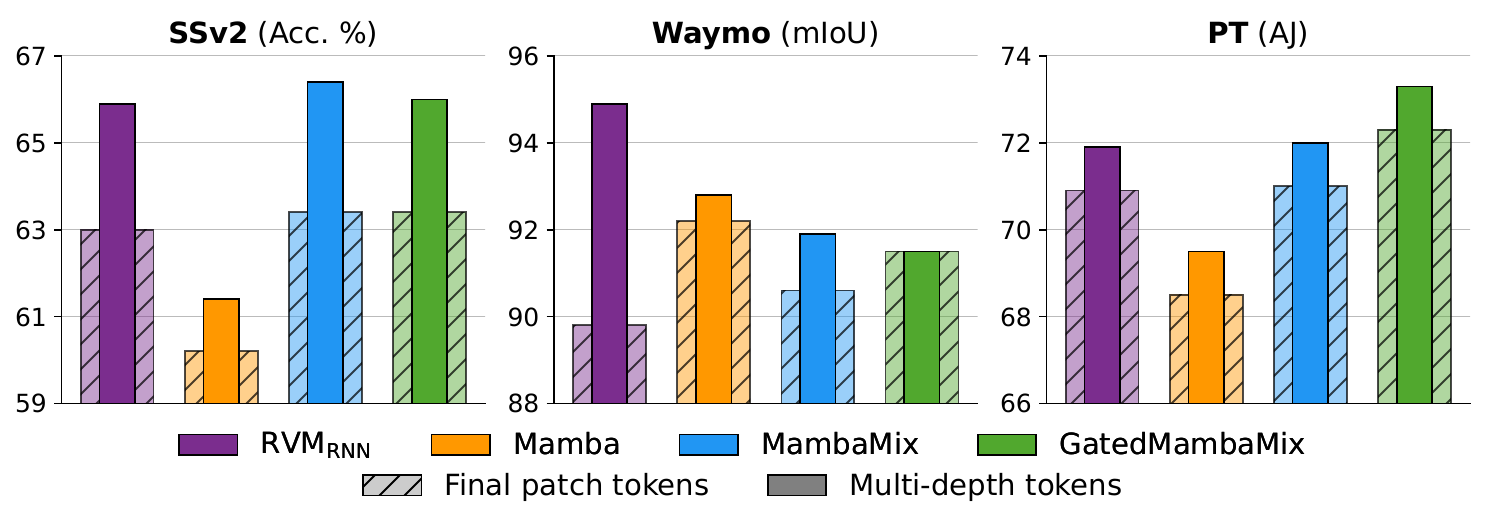}
    \caption{\textbf{Impact of Multi-depth Features} Using tokens from multiple DINOv3 depths (narrow solid bars) consistently improves or matches final-layer-only tokens (wide dashed bars) across all benchmarks and temporal architectures.}
    \label{fig:intermediate}
\end{figure*}

\input{tables/5_streaming_init_ablation}

\subsection{Results}

We now present results that address the two research questions posed in \cref{sec:intro}: whether image pre-training of the spatial encoder is competitive with video pre-training, and whether large-scale video pre-training is needed for the temporal module.

\mypara{Frame Encoder Comparison.}
\label{sec:ablations}
We pair GMMix with five encoders spanning different pre-training paradigms: DINOv3 and DINOv2~\cite{oquab_dinov2_2023, dinov3} (self-supervised), SigLIP2~\cite{tschannen2025siglip} (image-text contrastive), a ViT pre-trained on ImageNet-21K with supervision~\cite{pmlr-v139-touvron21a}, and the RVM encoder~\cite{rvm} (video pre-trained).
All encoders are frozen; only GMMix and the readout are trained from scratch under identical conditions.
\Cref{fig:encoder_comparison} reports the results.
On SSv2, DINOv3 (66.9) and DINOv2 (65.6) outperform the video pre-trained RVM encoder (62.7) by 4.2 and 2.9 points, respectively, while SigLIP2 (65.0) and even the supervised ViT-21K (61.6) remain competitive.
On Waymo, all image pre-trained encoders (85.0--85.5) surpass the RVM encoder (83.4), with the supervised ViT-21K (85.5) achieving the highest score.
On point tracking, the RVM encoder (71.1) leads, but all image encoders remain within 3.3 points.
On depth estimation, DINOv3 (0.089 AbsRel) and DINOv2 (0.090) substantially outperform the RVM encoder (0.122), and both SigLIP2 (0.107) and ViT-21K (0.114) also improve over it.
These results directly address our first research question: image foundation models, and even purely supervised encoders, provide spatial features that are competitive with or superior to those from video pre-training.

\input{tables/1_models_overview}

\mypara{Temporal Architecture Comparison.}
Having established that DINOv3 provides a strong frozen encoder, we compare four temporal architectures (\cref{tab:streaming}).
All DINOv3-based configurations significantly outperform RVM\,(frozen): on SSv2, DINOv3 + RVM\textsubscript{RNN} (67.1) and GMMix (66.9) exceed RVM\,(frozen) (46.9) by over 20 points; on Waymo, all DINOv3 variants (84.8--85.7) surpass RVM\,(frozen) (72.7) by at least 12 points; on ScanNet, DINOv3 models (0.087--0.096 AbsRel) roughly halve the error of RVM\,(frozen) (0.129).
While GMMix achieves the highest normalized average (99.4), no single architecture dominates across all tasks: RVM\textsubscript{RNN} leads on SSv2 (67.1) and Waymo (85.7), and MMix achieves the best depth estimation (0.087 AbsRel). 
Rotational error (RPE\textsubscript{rot}) is virtually identical across all models (0.08--0.09$^\circ$) and yields no discriminative signal.
This diversity supports the decoupled paradigm: a single frozen image encoder can serve as a shared spatial backbone for different lightweight temporal heads, each selected or specialized for a given downstream task.

\mypara{Multi-depth Feature Extraction.}
Since the image encoder is frozen, we investigate whether extracting features from multiple ViT depths can further improve performance.
Our default configuration feeds the temporal module with tokens from four DINOv3 depths (at 1/4, 2/4, 3/4, and 4/4 of the network), as well as \texttt{CLS}
and register tokens.
\Cref{fig:intermediate} compares this multi-depth setup against a baseline that uses only final-layer patch tokens.
Multi-depth features consistently improve performance across all temporal architectures and benchmarks.
On SSv2, gains range from 1.2 (Mamba) to 3.0 points (MambaMix); on Waymo, the RVM\textsubscript{RNN} variant improves by 5.1 mIoU (89.8 to 94.9).
On point tracking, all four architectures gain 1.0 AJ.
Because a frozen encoder retains useful spatial information distributed across its depth rather than consolidating it at the final layer, multi-depth extraction recovers complementary representations that a single output layer would miss.

\mypara{Temporal Module Transfer.}
The previous experiments show that temporal modules trained from scratch already perform well, but our second research question asks whether video pre-training of the temporal module provides additional value.
Without performing the computationally intensive pre-training ourselves, we leverage the pre-trained weights of RVM's sequential core and compare two initialization strategies for the temporal module on top of frozen DINOv3: training from scratch \vs initializing from RVM's pre-trained weights (\cref{tab:stream_rvm_init_ablation}).
Pre-training the temporal module before fine-tuning is beneficial in both settings: RVM itself improves consistently when initialized from pre-trained rather than random weights ({+}9.5 SSv2, {+}5.1 Waymo, {+}5.6 PT, {$-$}0.032 ScanNet, {$-$}20.85\,mm NuScenes), and DINOv3 similarly benefits despite the temporal module having been pre-trained with a different encoder ({+}1.3 SSv2, {+}1.1 Waymo, {+}4.9 PT, {$-$}0.003 ScanNet, {$-$}4.81\,mm NuScenes).
This positive transfer indicates that the temporal module captures dynamics that are at least partially encoder-agnostic, and that video pre-training of the temporal module is beneficial regardless of whether the spatial encoder matches the one used during pre-training.
Moreover, in our architecture, the vast majority of model parameters and compute reside in the vision encoder, while the temporal module remains lightweight. Together with the gains from fine-tuning an already pre-trained temporal module rather than using it frozen, this motivates a practical serving paradigm: a single shared frozen encoder across tasks, paired with small per-task temporal heads that are first pre-trained on video and then fine-tuned for their dedicated downstream task.
These findings answer our second research question and support a paradigm where a frozen image encoder is combined with a video pre-trained temporal module, decoupling spatial and temporal learning entirely.
Crucially, this decoupled design is far more efficient than end-to-end video pre-training: the spatial encoder requires no video data at all, and only the lightweight temporal module needs to be exposed to video sequences, drastically reducing both the computational cost and the volume of video data required for pre-training.

\begin{figure}[t]
    \centering
    \includegraphics[width=\linewidth]{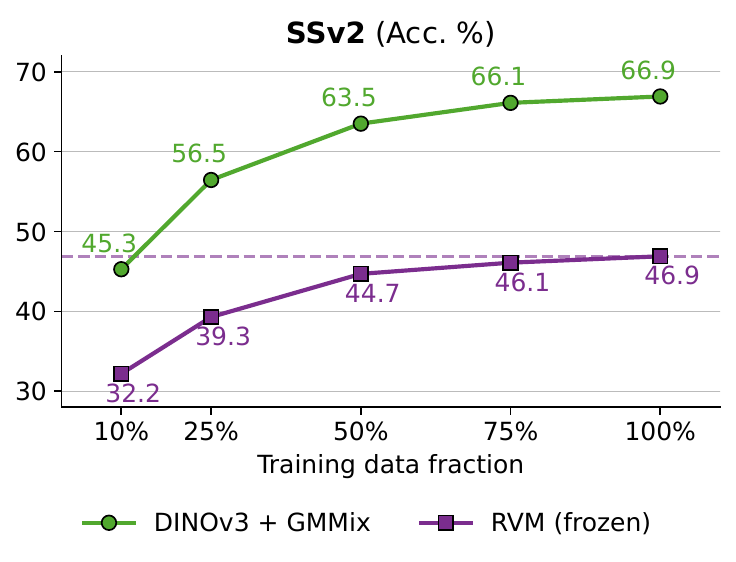}
    \caption{\textbf{Data Efficiency on SSv2.} DINOv3 + GMMix \vs frozen RVM trained on varying fractions of the SSv2 training set. Dashed line: frozen RVM at 100\%. DINOv3 + GMMix surpasses frozen RVM's full-data performance using less than 25\% of the training data.}
    \label{fig:data_efficiency}
\end{figure}

\mypara{Data Efficiency.}
We investigate how much downstream training data is needed by training DINOv3 + GMMix and frozen RVM on varying fractions of the SSv2 training set (\cref{fig:data_efficiency}).
At only 25\% of the data, DINOv3 + GMMix (56.5) already significantly surpasses frozen RVM trained on the full dataset (46.9).
The result suggests that a strong frozen image encoder provides sufficient spatial priors for the temporal module to learn effectively even from limited task-specific data.

\mypara{Comparison to Video Foundation Models.}
Finally, we compare our best model against established video foundation models (\cref{tab:general}).
All backbones are frozen; only a task-specific readout is trained.
For video foundation models (VideoMAE, V-JEPA, 4DS, RVM), the encoder already incorporates temporal information from end-to-end video pre-training.
Our model uses a frozen DINOv3 encoder with a GMMix temporal module trained from scratch on each downstream dataset, without any video pre-training.
Notably, frozen image encoders without any temporal module (DINOv3-L\textsubscript{dist}, DINOv2-L\textsubscript{dist}) already achieve competitive Waymo scores (78.8, 51.7) but fall short on temporally demanding tasks like SSv2 and point tracking, confirming that temporal modeling is necessary.
At ViT-L scale, our model (66.4 SSv2, 94.9 Waymo, 73.3 PT) is competitive with or exceeds all video pre-trained baselines, achieving the highest normalized average (99.1 vs.\ 89.3 for RVM-L).
At ViT-B scale, the same pattern holds: our model reaches 96.7 normalized average vs.\ 85.9 for RVM-B.
These results jointly confirm both research questions: image pre-training provides a spatial encoder that is competitive with video pre-training, and strong video understanding is achievable without large-scale video pre-training when the spatial encoder is sufficiently powerful.

%% file: tables/5_streaming_init_ablation.tex
\begin{table*}
\centering
\setlength{\tabcolsep}{7.5pt}
\begin{tabular}{lcllllll}
\toprule
\multirow{2}{*}{Model} & \multirow{2}{*}{Init} & SSv2 & Waymo & PT & ScanNet & NuScenes & NuScenes \\
& & Acc. ($\uparrow$\%) & mIoU ($\uparrow$) & AJ ($\uparrow$) & AbsRel ($\downarrow$) & RPE\textsubscript{tr} ($\downarrow$) & RPE\textsubscript{rot}, ($\downarrow$) \\
\midrule
RVM-L & Random           & 62.0           & 78.3           & 68.4           & 0.1237 & 38.66 & 0.10 \\
RVM-L & Pre-train              & 71.5\gain{9.5} & 83.4\gain{5.1} & 74.0\gain{5.6} & 0.0916\ngain{0.0321} & 17.81\ngain{20.85} & 0.05\ngain{0.05} \\
\midrule
DINOv3-L + RVM\textsubscript{RNN} & Random  & 67.1           & 85.7           & 63.7           & 0.0900 & 29.37 & 0.09 \\
DINOv3-L + RVM\textsubscript{RNN} & Pre-train & 68.4\gain{1.3} & 86.8\gain{1.1} & 68.6\gain{4.9} & 0.0866\ngain{0.0034} & 24.56\ngain{4.81} & 0.08\ngain{0.01} \\
\bottomrule
\end{tabular}
\caption{\textbf{Video Pre-training Transfer of the Temporal Module.} Init: temporal module initialization. Pre-train: temporal module initialized from RVM's pre-trained weights. Pre-training the temporal module before fine-tuning consistently improves performance in both settings: when used with the original RVM encoder and when transferred to a different encoder (DINOv3), indicating that learned temporal dynamics are partially encoder-agnostic. NuScenes RPE\textsubscript{tr} is in mm, RPE\textsubscript{rot} is in $^\circ$.}
\label{tab:stream_rvm_init_ablation}
\end{table*}

%% file: tables/1_models_overview.tex
\begin{table*}
    \centering
    \begin{tabular}{lccccc}
    \toprule
    \multirow{2}{*}{Model} & \multirow{2}{*}{Pre-train} & SSv2 & Waymo & PT & Norm. Avg \\
    & & Acc. ($\uparrow$\%) & mIoU ($\uparrow$) & AJ ($\uparrow$) & ($\uparrow$) \\
    \midrule
    DINOv3-L\textsubscript{dist} & Image & 55.9 & 78.8          & 38.6          & 72.7  \\
    DINOv2-L\textsubscript{dist} & Image & 52.2 & 51.7           & 42.4          & 63.1  \\
    VideoMAE-L           & Video & 62.7          & 74.9          & 70.5          & 88.9  \\
    V-JEPA-L             & Video & 66.0          & 73.3          & 67.1          & 88.5  \\
    4DS-L                & Video & 57.6          & 75.9          & n/a           & n/a  \\
    RVM-L                & Video & \textbf{66.7} & 73.2          & 68.1       & 89.3  \\
    \textbf{DINOv3-L + GMMix} & Image & 66.4   & \textbf{94.9} & \textbf{73.3} & \textbf{99.1}  \\
    \midrule
    DINOv3-B\textsubscript{dist} & Image & 50.6 & 78.7         & 41.3          & 71.3  \\
    VideoMAE-B           & Video & 52.3          & 73.1          & 71.3             & 83.5  \\
    4DS-B                & Video & 49.6          & 72.7          & n/a           & n/a  \\
    4DS-B\textsubscript{dist} & Video & 60.3  & 76.3          & 70.4          & 88.2  \\
    RVM-B                & Video & \textbf{61.4} & 71.1          & 68.1          & 85.9  \\
    \textbf{DINOv3-B + GMMix} & Image & 60.7   & \textbf{93.9}          & \textbf{75.1} & \textbf{96.7} \\
    \bottomrule
    \end{tabular}
    
    \caption{\textbf{Comparison to Video Foundation Models.} All vision encoders are frozen, only a task-specific readout head is trained for all models. Our model additionally fine-tunes a lightweight temporal module from scratch, without any large-scale video pre-training. n/a: checkpoint not publicly available.}
    \label{tab:general}
\end{table*}

%% file: sec/5_conclusion.tex
\section{Conclusion}
\label{sec:conclusion}

We investigate whether end-to-end video pre-training is necessary for strong video understanding, or whether spatial and temporal learning can be effectively decoupled.
Our experiments demonstrate that a frozen DINOv3 image encoder paired with a lightweight recurrent temporal module, trained from scratch, matches the performance of RVM, a model pre-trained end-to-end on video, under the same evaluation protocol.
This result holds across multiple temporal module architectures, from gated transformer cores to Mamba-based variants, indicating that the quality of the spatial encoder is the dominant factor rather than the specific temporal design.

Our findings have practical implications: rather than investing in end-to-end video pre-training from scratch, practitioners can leverage existing pre-trained image encoders and perform a lightweight video pre-training of the temporal module instead. The transferability of temporal modules across encoders further suggests that spatial and temporal representations are naturally separable, opening the door to modular video model design where components can be developed and improved independently.

Since no dominant streaming video architecture has yet emerged, this modular design offers flexibility in performance, latency, and memory trade-offs: the temporal module can be freely chosen to fit a target application.

\paragraph{Limitations.}
In this preliminary work, we did not take the step of pre-training the temporal module on video data to fully demonstrate the advantages of decoupled pre-training. This is deferred to future work, and the findings of this paper provide sufficient grounds to make the required investments. Such a future study should also include model size beyond Base and Large, more families of image and video encoders and pre-training objectives, and a more comprehensive comparison to alternative video methods.

%% file: sec/6_acknowledgements.tex
\section*{Acknowledgements}

This work was funded by the European Union, under grant agreement 101076810 (project MODI). 
We also acknowledge the Dutch national e-infrastructure with the support of the SURF Cooperative, grant agreement no. EINF-16686, financed by the Dutch Research Council (NWO), for the availability of high-performance computing resources and support.

%% file: sec/X_suppl.tex
\clearpage
\setcounter{page}{1}
\maketitlesupplementary

The offline evaluation protocol, including readout architectures and training procedures, follows~\cite{rvm,4ds}. In this supplementary material, we explain the streaming evaluation, which reflects the real-world scenario of receiving video frames one by one and processing them as soon as they arrive. In this setting, the readout operates on a single frame's tokens at each time step, and all temporal context must reside in the recurrent state of the temporal module.

\section{Streaming Tasks}
\label{sec:suppl_tasks}

\begin{table*}[t]
\centering
\caption{\textbf{Streaming task overview.} Training and evaluation setup for each downstream task.}
\label{tab:suppl_tasks}
\begin{tabular}{l >{\raggedright\arraybackslash}p{3.8cm} l l l}
\toprule
Task & Dataset & Train Loss & Eval Metric & \begin{tabular}[c]{@{}c@{}}Readout \\ (dim, heads)\end{tabular} \\
\midrule
Action Recognition & SSv2~\cite{ssv2} & Cross-entropy & Top-1 Acc.\ (\%) $\uparrow$ & (768, 12) \\
Object Tracking & Waymo Open~\cite{waymo} & GIoU + L1 & mIoU $\uparrow$ & (1024, 4) \\
Point Tracking & Kubric MOVi-E~\cite{greff2022kubric} (train),\newline PT~\cite{perceptiontest} (eval) & Huber + BCE & Avg.\ Jaccard $\uparrow$ & (1024, 8) \\
Depth Estimation & ScanNet~\cite{scannet} & Log-space L2 & AbsRel $\downarrow$ & (1024, 16) \\
Pose Estimation & NuScenes~\cite{nuscenes} & L1 (learned balancing) & RPE\textsubscript{tr}\,(mm) $\downarrow$ & (1024, MLP) \\
\bottomrule
\end{tabular}
\end{table*}

\Cref{tab:suppl_tasks} provides an overview of each task, including the training dataset, loss function, evaluation metric, and readout head parameters.
The readout heads are based on the cross-attention architecture from~\cite{rvm}, adapted to operate on single-frame tokens ($N$ tokens per frame) instead of the full spatio-temporal sequence ($T \times N$ tokens). The encoder is always frozen; only the temporal module and readout head receive gradients. In the case of pre-trained RVM models, only the readout is trained.

\subsection{Action Recognition (SSv2)}

The offline readout attends to all $T \times N$ spatio-temporal tokens with learned temporal positional embeddings. In the streaming setting, the readout instead attends to the current frame's $N$ tokens only (12 heads, $d{=}768$), without temporal positional embeddings, since temporal context is captured entirely by the recurrent state. A single learned query is projected to 174 classes. The final prediction $\hat{y}_T$ from the last frame is used for evaluation. The loss is standard cross-entropy.

\subsection{Object Tracking (Waymo)}

Following the offline protocol~\cite{rvm,4ds}, the initial bounding box $[c_x, c_y, w, h]$ is encoded via 16 Fourier frequencies and processed through an MLP (hidden dim 512) to produce a single query token. The difference is in the readout: instead of attending to all $T \times N$ tokens and predicting all frames at once, the streaming readout processes one frame at a time. At each frame $t$, the query attends to the current frame's $N$ tokens via cross-attention ($d{=}1024$, 4 heads), and the updated query is projected through an MLP to 4 bounding box coordinates. The refined query is then carried over to frame $t{+}1$, acting as an evolving tracker state alongside the temporal module's recurrent state. The loss combines GIoU (weight 2.0) and L1 (weight 5.0) on the predicted coordinates.

\subsection{Point Tracking (Perception Test)}

Models are trained on the synthetic Kubric MOVi-E dataset~\cite{greff2022kubric} and evaluated on real videos from Perception Test, constituting a synthetic-to-real transfer setting. Each sample uses 64 query points, initialized with their ground-truth $(x, y)$ position at the first frame. Each query point is encoded via 16 Fourier frequencies and processed through an MLP ($2 \times 512$ hidden units), then linearly projected to $d{=}1024$.

The offline readout replicates each query 8 times with learnable temporal embeddings, and each of the 8 queries predicts 2 consecutive frames while attending to the full $T \times N$ spatio-temporal tokens. In the streaming setting, the readout instead uses a single query per point, without temporal embeddings: at each frame, the query attends to the current frame's $N$ tokens via cross-attention ($d{=}1024$, 8 heads) and directly predicts position $(x, y)$, visibility, and uncertainty for that frame. The loss combines Huber loss ($\delta{=}0.05$, weight 100.0) on positions (visible points only), binary cross-entropy on visibility (weight 0.1), and binary cross-entropy on uncertainty (weight 0.1).

\subsection{Depth Estimation (ScanNet)}

The offline readout uses spatio-temporal $2 \times 8 \times 8$ patches as queries over all frames, while the streaming readout uses $\frac{H}{8} \times \frac{W}{8}$ spatial patches per frame. Since we use a fixed input resolution of $224 \times 224$, this gives $28 \times 28 = 784$ learned queries per frame. Each query attends to the current frame's $N$ tokens via cross-attention ($d{=}1024$, 16 heads) and predicts $8 \times 8 = 64$ depth values for its patch, which are rearranged to produce a full-resolution $224 \times 224$ depth map.

\subsection{Camera Pose Estimation (NuScenes)}

At each frame $t$, the readout aggregates the $N$ spatial tokens via mean pooling and passes the pooled vector through an MLP (LayerNorm, Linear $d{\to}512$, GELU, Linear $512 {\to} 9$) to predict a 9-dimensional pose delta: 3 values for translation $(dx, dy, dz)$ and a 6-dimensional rotation representation~\cite{zhou2019continuity}, which avoids the discontinuities of quaternions. Predictions are frame-to-frame deltas (the pose change from frame $t{-}1$ to frame $t$).

The loss uses learnable translation/rotation balancing following Kendall \& Cipolla~\cite{kendall2018multi}:
\begin{equation}
    \mathcal{L} = \mathcal{L}_{\mathrm{trans}}\, e^{-s_t} + s_t + \mathcal{L}_{\mathrm{rot}}\, e^{-s_r} + s_r,
\end{equation}
where $\mathcal{L}_{\mathrm{trans}}$ and $\mathcal{L}_{\mathrm{rot}}$ are the L1 losses on the translation and 6D rotation components, respectively, and $s_t, s_r$ are learnable log-variance parameters that automatically balance the two terms.

\section{Training Settings}
\label{sec:suppl_training}

All tasks share the same training configuration unless stated otherwise. We use AdamW~\cite{loshchilov2017adamw} with $(\beta_1, \beta_2) = (0.9, 0.999)$, weight decay $10^{-4}$, a cosine learning rate schedule decaying to $\eta_{\min} = 10^{-7}$, and linear warmup. Training uses mixed precision (bf16). Our protocol is based on 4DS~\cite{4ds}, which used 40K steps for frozen training (only the readout trainable) and 80K steps for fine-tuning. We train for 40K steps in the frozen regime and 100K steps in the RNN fine-tuning regime. Streaming experiments use a dedicated protocol of 20K training steps. The three regimes differ in which components are trained and in the warmup length, as summarized in \cref{tab:suppl_regimes}.

\begin{table}[t]
\centering
\small
\caption{\textbf{Training regimes.} All regimes use AdamW, cosine schedule to $\eta_{\min}=10^{-7}$, and bf16 mixed precision.}
\label{tab:suppl_regimes}
\begin{tabular}{lccc}
\toprule
Regime & Trainable & Train steps & Warmup \\
\midrule
Frozen              & readout only           & 40{,}000  & 1{,}000 \\
RNN fine-tuning     & RNN + readout & 100{,}000 & 5{,}000 \\
Streaming           & RNN + readout & 20{,}000  & 1{,}000 \\
\bottomrule
\end{tabular}
\end{table}

The only setting that varies across tasks and models is the peak learning rate, reported in \cref{tab:suppl_lr_offline,tab:suppl_lr_streaming}.

\begin{table}[t]
\centering
\small
\caption{\textbf{Peak learning rates (offline).} Frozen and RNN fine-tuning regimes. Dv3: DINOv3.}
\label{tab:suppl_lr_offline}
\begin{tabular}{llccc}
\toprule
Regime & Model & SSv2 & Waymo & Kubric \\
\midrule
\multirow{2}{*}{Frozen} & RVM (frozen)            & 3e-4   & 3e-4 & 1e-4 \\
                        & Dv3 (no RNN)    & 1e-4   & 1e-4   & 5e-5 \\
\midrule
\multirow{4}{*}{\shortstack{RNN\\fine-tune}} & Dv3 + RVM\textsubscript{RNN}  & 1e-4 & 1e-4 & 5e-5 \\
                               & Dv3 + Mamba                    & 1e-4 & 1e-4 & 5e-5 \\
                               & Dv3 + MMix                 & 1e-4 & 1e-4 & 5e-5 \\
                               & Dv3 + GMMix                    & 1e-4 & 1e-4 & 5e-5 \\
\bottomrule
\end{tabular}
\end{table}

\begin{table}[t]
\centering
\small
\setlength{\tabcolsep}{3pt}
\caption{\textbf{Peak learning rates (streaming).} All models trained for 20K steps with 1K warmup. Dv3: DINOv3.}
\label{tab:suppl_lr_streaming}
\begin{tabular}{lccccc}
\toprule
Model & SSv2 & Waymo & Kubric & ScanNet & NuScenes \\
\midrule
RVM (frozen)                        & 1e-4 & 1e-4 & 3e-4 & 1e-4 & 1e-3 \\
Dv3 + RVM\textsubscript{RNN}     & 1e-4   & 5e-5   & 1e-4   & 5e-5   & 3e-4 \\
Dv3 + Mamba                       & 1e-4 & 5e-5   & 1e-4 & 5e-5 & 3e-4 \\
Dv3 + MMix                    & 1e-4 & 5e-5 & 1e-4 & 5e-5 & 3e-4 \\
Dv3 + GMMix                       & 1e-4 & 5e-5 & 1e-4 & 5e-5 & 3e-4 \\
\bottomrule
\end{tabular}
\end{table}

\section{Evaluation Metrics}
\label{sec:suppl_metrics}

\mypara{Top-1 accuracy (SSv2).}
Standard classification accuracy on the validation set.

\mypara{mIoU (Waymo).}
Mean Intersection over Union between predicted and ground-truth bounding boxes, averaged over all objects and frames.

\mypara{Average Jaccard (PT).}
Following the Perception Test benchmark~\cite{perceptiontest}, AJ is defined as the average of Jaccard values at position thresholds of 1, 2, 4, 8, and 16 pixels, where a point is considered correctly tracked if its predicted position is within the threshold and its visibility is correctly predicted.

\mypara{AbsRel (ScanNet).}
Absolute relative error:
\begin{equation}
    \text{AbsRel} = \frac{1}{|\mathcal{P}|} \sum_{p \in \mathcal{P}} \frac{|d_p - \hat{d}_p|}{d_p},
\end{equation}
where $d_p$ and $\hat{d}_p$ are the ground-truth and predicted depth at pixel $p$, and $\mathcal{P}$ is the set of valid pixels.

\mypara{RPE\textsubscript{tr} and RPE\textsubscript{rot} (NuScenes).}
Translational (mm) and rotational (degrees) components of the relative pose error between consecutive frames.

\mypara{Normalized average.}
Each score is divided by the column-best across all models in the table, and the ratios are averaged.
For metrics where lower is better (AbsRel, RPE\textsubscript{tr}), we use (column-best / score) instead of (score / column-best).